\documentclass[sigconf, nonacm, 9pt]{acmart}

\usepackage{graphicx}
\usepackage{todonotes}
\usepackage{mathtools}
\usepackage{breqn}
\usepackage{algorithm}
\usepackage[noend]{algpseudocode}
\usepackage{amsmath}
\usepackage{subcaption}
\usepackage{tabularx}
\usepackage{hyphenat}
\usepackage{fancyvrb}
\usepackage{bm}

\acmConference[]{}{}{}{}

\newcommand{\squeeze}{\vspace{-0.3cm}}

\setcopyright{none}

\begin{document}

\title[Differentiable Agent-Based Simulation for Gradient-Guided Simulation-Based Optimization]{Differentiable Agent-Based Simulation for\\Gradient-Guided Simulation-Based Optimization}

\author{Philipp Andelfinger}
\affiliation{
   \institution{University of Rostock}
  \department{Institute for Visual and Analytic Computing}
  \city{Rostock}
  \postcode{18051}
  \country{Germany}
}


\begin{abstract}
Simulation-based optimization using agent-based models is typically carried out under the assumption that the gradient describing the sensitivity of the simulation output to the input cannot be evaluated directly.
To still apply gradient-based optimization methods, which efficiently steer the optimization towards a local optimum, gradient estimation methods can be employed.
However, many simulation runs are needed to obtain accurate estimates if the input dimension is large.
Automatic differentiation (AD) is a family of techniques to compute gradients of general programs directly.
Here, we explore the use of AD in the context of time-driven agent-based simulations.
By substituting common discrete model elements such as conditional branching with smooth approximations, we obtain gradient information across discontinuities in the model logic.
On the example of microscopic traffic models and an epidemics model, we study the fidelity and overhead of the differentiable models, as well as the convergence speed and solution quality achieved by gradient-based optimization compared to gradient-free methods.
In traffic signal timing optimization problems with high input dimension, the gradient-based methods exhibit substantially superior performance.
Finally, we demonstrate that the approach enables gradient-based training of neural network-controlled simulation entities embedded in the model logic.
\end{abstract}

\maketitle

\section{Introduction}

Simulation-based optimization comprises methods to determine a simulation input parameter combination that minimizes or maximizes an output statistic~\cite{carson1997simulation,hong2009brief}, with applications in a vast array of domains such as supply chain management~\cite{jung2004simulation}, transportation~\cite{osorio2015computationally}, building planning~\cite{yigit2018simulation}, and health care~\cite{yousefi2020simulation}.
The problem can be viewed as a special case of mathematical optimization in which an evaluation of the objective function is reflected by the execution of one or more simulation runs.
Many mathematical optimization methods evaluate not only the objective function itself but also its partial derivatives to inform the choice of the next candidate solution.
Given a suitable initial guess, gradient-based methods efficiently steer the optimization towards a local optimum, with provable convergence under certain conditions~\cite{ruder2016overview}.

In contrast, simulation-based optimization using agent-based models usually relies either on surrogate models, which typically abandon the individual-based level of detail of the original model~\cite{anirudh2020accurate}, or on gradient-free methods such as genetic algorithms~\cite{calvez2005automatic}.
While gradient-free simulation-based optimization is a time-tested approach, the hypothesis underlying the present paper is that the targeted local search carried out by gradient-based methods may achieve faster convergence or higher-quality solutions for certain agent-based models.
An existing method to obtain gradients is Infinitesimal Perturbation Analysis (IPA)~\cite{ho1983new}, which the literature applies by determining derivative expressions by a manual model analysis (e.g.,~\cite{howell2006simulation,chen2010perturbation,geng2012multi}), limiting its applicability to relatively simple models.
Alternatively, gradients can be estimated based on finite differences.
However, the number of required simulation runs grows linearly with the input dimension, rendering this approach prohibitively expensive in non-trivial applications.

\begin{figure}[t!]
    \centering
\begin{minipage}[b]{.45\textwidth}
    \centering
    \hspace{-0.15cm} 
    \includegraphics[width=\textwidth]{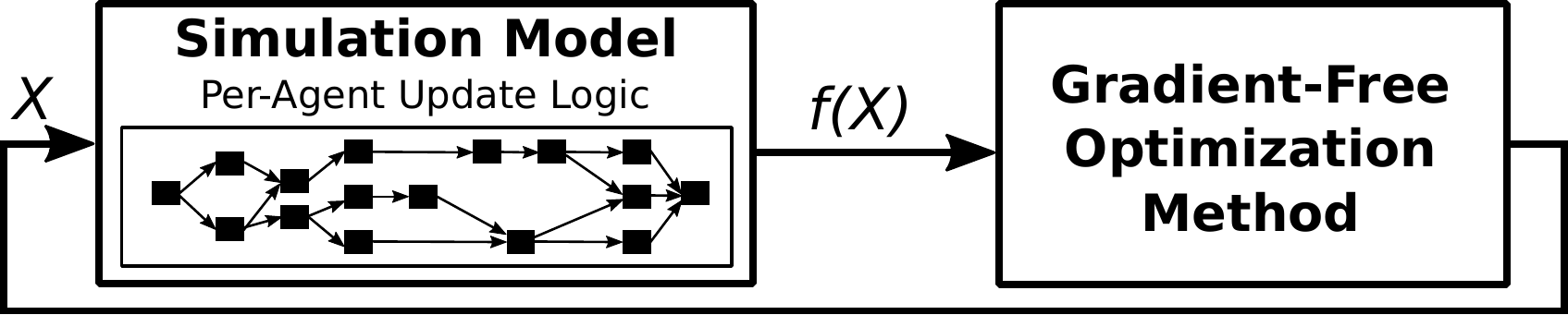}
    \subcaption{Gradient-free simulation-based optimization.}
    \vspace{0.2cm}
\end{minipage}
\begin{minipage}[b]{.45\textwidth}
    \centering
    \includegraphics[width=\textwidth]{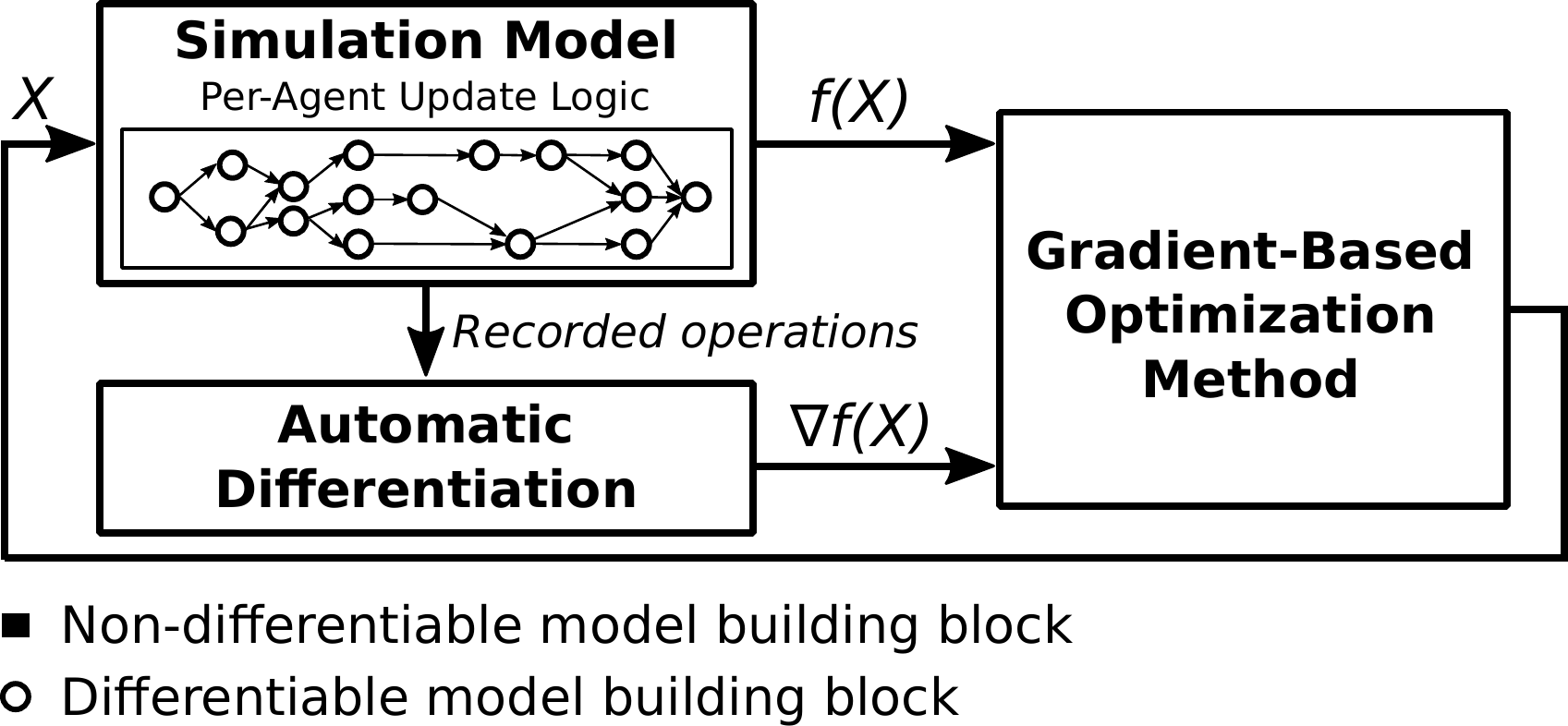}
    \subcaption{Gradient-based simulation-based optimization.}
\end{minipage}
    \vspace{-0.1cm}
	\caption{Our approach: by substituting agent model elements with differentiable counterparts, automatic differentiation can be used to enable gradient-based optimization.}
    \label{fig:illustration}
    \vspace{-0.20cm}
\end{figure}

A similar problem, as well as an efficient solution, exists in the field of deep learning, where the ability to optimize neural networks with millions of parameters within tolerable timeframes rests on gradient information determined using the backpropagation algorithm~\cite{rumelhart1986learning}.
Backpropagation is a special case of automatic differentiation, a family of methods to compute derivatives of computer programs written in general-purpose programming languages~\cite{margossian2019review}.

In this paper, we explore the use of automatic differentiation for gradient-based optimization of agent-based simulations.
The main obstacle towards this goal is given by discontinuous model elements, which are considered among the constituent features of agent-based models~\cite{bonabeau2002agent}.
To enable differentiation across such elements, we propose substitutes constructed from known smooth approximations of basic operations (cf.~Fig.~\ref{fig:illustration}).
In contrast to classical surrogate modeling approaches, our approach retains the per-agent logic of the original model.
The resulting agent-based models are differentiable regarding some or all model aspects, depending on the degree to which discontinuous elements are substituted.
We refer to this approach as \emph{differentiable agent-based simulation}.

To evaluate the approach, we implement models from the transportation and epidemics domains in a differentiable fashion and study 1.~the fidelity of the results as compared to purely discrete reference models, 2.~the overhead introduced by relying on smooth model building blocks, and 3.~the relative convergence behavior and solution quality in simulation-based optimization problems as compared to gradient-free methods.
To further showcase the potential of the approach, we extend the traffic simulation model by embedding neural network-controlled traffic signals in the differentiable model logic, which enables their training using gradient-based methods.

Our main contributions are as follows:
\begin{itemize}
\item \textbf{Differentiable agent-based simulation}, i.e., the construction of agent-based models from differentiable building blocks to enable automatic differentiation and gradient-based optimization. An initial set of building blocks to construct differentiable model implementations is presented.
\item \textbf{Differentiable model implementations} based on well-known models from the literature to demonstrate the approach. We measure the fidelity and performance compared to discrete reference implementations.
\item \textbf{Comparison of convergence and solution quality} in a simulation-based traffic signal timing optimization using gradient-free and gradient-based methods.
\item \textbf{Embedding and training of neural network-controlled entities}, demonstrating their training based on simulation gradient information on the example of dynamic traffic lights.
\squeeze
\end{itemize}

The remainder of the paper is structured as follows:
In Section~\ref{sec:automatic_differentiation}, we briefly introduce automatic differentiation, which forms the basis for our approach.
In Section~\ref{sec:differentiable_agent-based_simulation}, the concept of differentiable agent-based simulation is introduced and building blocks are provided to construct differentiable models.
In Section~\ref{sec:model_implementation}, we describe the differentiable models we implemented to demonstrate the approach.
In Section~\ref{sec:experiments}, experiment results are presented to evaluate the performance and fidelity of the model implementations as well as the benefits of our approach in simulation-based optimization problems. 
In Section~\ref{sec:limitations_and_research_directions}, we discuss limitations of the approach and various directions for future research.
Section~\ref{sec:related_work} describes related work.
Section~\ref{sec:conclusions} summarizes our results and concludes the paper.

\section{Automatic Differentiation}
\label{sec:automatic_differentiation}
In this section, we give a brief overview of Automatic Differentiation (AD), which comprises techniques to computationally determine gradients of computer programs~\cite{margossian2019review}.
In contrast to finite differences, AD determines exact partial derivatives for an arbitrary number of either input or output variables from a single program execution.
In comparison to symbolic differentiation as implemented in computer algebra systems, AD avoids representing the frequently long derivative expressions explicitly~\cite{baydin2017automatic}.

AD computes derivatives based on the chain rule from differential calculus.
In the \emph{forward} mode, intermediate results of differentiating the computational operations w.r.t.~one of the inputs are carried along during the program's execution.
At termination, the partial derivatives of all output variables w.r.t.~one input variable have been computed.
Thus, given $n$ input variables, $n$ passes are required.
Conversely, \emph{reverse-mode} AD computes the derivatives of one output variable w.r.t.~arbitrarily many inputs in a single pass.
Given our use case of simulation-based optimization, where we expect the input dimension to be larger than the output dimension, the remainder of the paper will rely on reverse-mode AD.

\begin{figure}[t]
\hspace{-1.0cm}
\begin{tikzpicture}[node distance={15mm}, thick, main/.style = {draw, circle, text width=0.45cm, align=center}]
\node[main,draw=none,label=left:{$v_1$}] (1) {$\boldsymbol{x}$};
\node[main,label=right:{$v_3$}] (3) [right of=1] {$\boldsymbol{(~)^2}$};
\node[main,label=above:{$v_4$}] (4) [below right of=3] {$\boldsymbol{\times}$};
\node[main,draw=none,label=left:{$v_2$}] (2) [left of=4] {$\boldsymbol{y}$};
\node[main,label=above:{$v_5$}] (5) [right of=4] {\textbf{sin}};
\draw[->] (1) -- (3);
\draw[->] (3) -- (4);
\draw[->] (2) -- (4);
\draw[->] (4) -- (5);
\end{tikzpicture}
\caption{Simple example function \bm{$\sin(x^2y)$}. Automatic differentiation propagates values reflecting the sensitivities of the intermediate results w.r.t.~the inputs along the nodes \bm{{$v_i$}}.}
\label{fig:ad_example}
\squeeze
\end{figure}
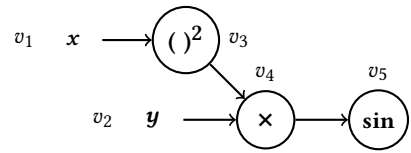

During the execution of the program, the computational operations and intermediate results are recorded in a graph.
At termination, the graph is traversed in reverse, starting from the output.
The chain rule is applied at each operation to update the intermediate derivative calculation.
When arriving at an input variable, the computed value is the partial derivative of the simulation output w.r.t.~the respective input.
We give a brief example of reverse-mode AD of a program implementing the function $f(x, y) = \sin(x^2y)$ (cf.~Figure~\ref{fig:ad_example}).
The intermediate results during execution (denoted by $v_i$) and during reverse-mode AD (denoted by $\bar{v}_i$) are as follows:\vspace{0.075cm}
{\noindent
\begin{tabular*}{\linewidth}{p{3.5cm}l}
$v_1 = x$                    & $\bar{v}_5 = 1$\\
$v_2 = y$                    & $\bar{v}_4 = \frac{\partial v_5}{\partial v_4} \bar{v}_5 = \cos(v_4) 1 = \cos(x^2y)$\\
$v_3 = {v_1}^2 = x^2$        & $\bar{v}_3 = \frac{\partial v_4}{\partial v_3} \bar{v}_4 = v_2 \bar{v}_4 = y \cos(x^2y)$\\
$v_4 = {v_3}v_2 = x^2y$      & $\bar{v}_2 = \frac{\partial v_4}{\partial v_2} \bar{v}_4 = v_3 \bar{v}_4 = x^2 \cos(x^2y)$\\
$v_5 = \sin(v_4) = \sin(x^2y)$ & $\bar{v}_1 = \frac{\partial v_3}{\partial v_1} \bar{v}_3 = 2v_1 \bar{v}_3 = 2xy\cos(x^2y)$
\vspace{0.075cm}
\end{tabular*}
}
The partial derivatives of the function are $\frac{\partial f}{\partial{x}} = \bar{v}_1 = 2xy\cos(x^2y)$ and $\frac{\partial f}{\partial y} = \bar{v}_2 = x^2\cos(x^2y)$.

Mature implementations of AD are available for programming languages such as C and C++~\cite{hogan2014fast,griewank1996algorithm}, Java~\cite{slucsanschi2016adijac}, and Julia~\cite{innes2019zygote}.
Modern AD tools rely on expression templates~\cite{hogan2014fast} or source-to-source transformation~\cite{innes2019zygote} to generate efficient differentiation code.

The backpropagation algorithm used to train neural networks is a special case of AD~\cite{baydin2017automatic}, where the computation graph is the sequence of multiplications, additions and activation function invocations reflecting a forward pass through the network.
Outside of machine learning, AD has been applied for solving differential equation~\cite{rackauckas2018comparison}, in physics applications~\cite{hu2019difftaichi}, and for optimal control~\cite{andersson2012casadi}.

There is a growing interest in expressing a broader range of general algorithms in a differentiable manner (e.g.~\cite{shaikhha2018efficient,cuturi2019differentiable,grover2019stochastic}).
Apart from enabling the gradient-based optimization of program parameters, differentiable programs can be directly integrated into network training pipelines, enabling gradient information to propagate through a combined computation graph comprised of a neural network and application-specific algorithms~\cite{de2018end,degrave2019differentiable}.

%

\section{Differentiable Agent-Based Simulation}
\label{sec:differentiable_agent-based_simulation}

From a computational perspective, a simulation model implementation applies a sequence of logical and arithmetic operations to the input variables to produce a set of output variables.
Our goal is to extract derivatives from model executions that can guide gradient-based optimization methods, which requires the operations to be differentiable and to yield finite and non-zero derivatives.
However, typical agent-based models must be expected to contain discontinuous elements: viewed as functions on the reals, model elements such as conditional branching and discrete state transitions may be non-differentiable in certain points or regions of their domain, and may carry unhelpful zero-valued derivatives in others.
The approach of differentiable agent-based simulation involves the construction of model implementations from differentiable building blocks that act as substitutes for non-differentiable model elements.

As a basic example, consider the following C code fragment containing a conditional statement:
\begin{Verbatim}[fontsize=\small]
bool f(double x, double x0) {
    if(x >= x0) return true;
    return false;
}
\end{Verbatim}

\begin{figure}[b]
\squeeze
\centering
\includegraphics[width=0.49\textwidth]{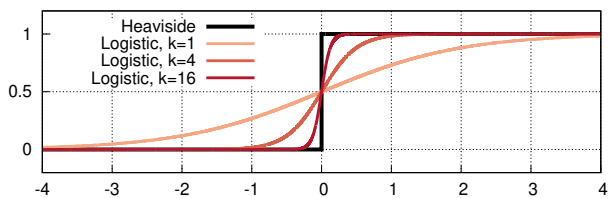}
\caption{Smooth approximation of the Heaviside step function using the logistic function.}
\label{fig:step_vs_logistic_function}
\end{figure}

For $x_0 = 0$, the function behaves like the Heaviside step function (cf.~Fig.~\ref{fig:step_vs_logistic_function}), the derivative of which is zero-valued at $x \ne 0$, and infinite at $x = 0$.
A well-known smooth approximation~\cite{kyurkchiev2015sigmoid} is given by the logistic function:
$\text{l}_k(x) = (1 + e^{-k(x - x_0)})^{-1}$, where k determines the steepness of the curve, and $x_0$ shifts the curve along the x axis.
In Fig~\ref{fig:step_vs_logistic_function}, we also show the logistic function with $x_0 = 0$, varying $k$.
When increasing $k$, the logistic function becomes a closer and closer approximation of the Heaviside step function.
To make our example C function amenable to automatic differentiation, we can simply substitute its body with \verb!return logistic(x, x0);!

By varying $x_0$, we can now approximate expressions of the form $x \ge x_0$.
In the remainder of the paper, we refer to this basic building block as \emph{smooth threshold}.
In the remainder of this section, we describe a number of more complex building blocks for agent-based simulations.
This bottom-up approach is similar in spirit to the construction of reversible programs~\cite{toffoli1980reversible,perumalla1999source}, which has also found applications in the simulation realm (e.g.,~\cite{yoginath2009reversible}). 
We emphasize that the building blocks described in the following rely on well-known continuous approximations and that the list is far from complete; our intention is to gather a basic list of constructs as a starting point for model development.

\subsection{Conditional Execution and Branching}
A challenge for automatic differentiation is given by control flow that depends on the program input.
Consider the code \verb!if(x >= x0)! \verb!y = c; else y = d;! where x is the input, x0, c and d are constants, and y is the output.
During each execution, the control flow covers exactly one of the branches.
Thus, y is always assigned a constant, and $\frac{dy}{dx}$, i.e., the sensitivity of y to changes in x, evaluates to 0.
By expressing the control flow using a smooth approximation (as shown, e.g., in~\cite{hernandez2019differentiable}), we can extract a derivative across both branches: \verb!double z = logistic(x, x0);!
\verb!y = z * c + (1 - z) * d;!
We refer to this simple pattern as \emph{smooth branching}.

\subsection{Iteration}
The core of a typical time-driven agent-based simulation is a nested loop comprised of an outer loop that iterates across the time steps and an inner loop that iterates across the agents.
Assuming that the number of time steps and agents are both constants, the loop represents static, i.e., input-independent, control flow, which requires no further preparation for automatic differentiation.

Input-dependent loop conditions can be transformed into guards if an upper bound for the number of iterations is known.
For example, variations in the number of agents can be implemented by masking the behavior of non-existent agents using \emph{smooth threshold}.
We give the example of incrementing an attribute by a constant for two alive agents, while the operation is masked for a third agent:
\begin{Verbatim}[fontsize=\small]
int max_num_agents = 3;
double alive[] = { 1.0, 1.0, 0.0 };
for(int aid = 0; aid < max_num_agents; ad++)
    agent[aid].attr += c * alive[agent_id];
\end{Verbatim}

\subsection{Selection of Interaction Partners}
\label{sec:differentiable_agent-based_simulation:subsec:selection_of_interaction_partners}
The selection of interaction partners based on their attributes is one of the fundamental primitives in agent-based simulations.
A straightforward differentiable solution to neighbor detection applies the principle described in the previous subsection: each agent iterates across all other agents, masking interactions with those for which a certain condition does not hold.

An example is given by the traffic simulation model of Section~\ref{sec:model_implementation}: each vehicle chooses its acceleration based on attributes of the closest vehicle ahead.
The selection of the minimum distance can be achieved by iteratively applying a smooth approximation of the minimum function: $-\log\sum_i{e^{-x_i}}$~\cite{cook2011basic}.
Once the distance to the closest vehicle has been determined, additional attributes are required to carry out the interaction.
To support the retrieval of additional attributes, we construct a \emph{select by attribute} building block, which selects an agent's specified ``target'' attribute based on the known value of a ``reference'' attribute.
This is achieved by iterating across all agents' reference attributes, adding the target attribute only if the reference attribute is sufficiently close to the known value, based on an \verb!in_range! function. The \verb!in_range! function then applies \emph{smooth threshold} with bounds \verb!-eps! and \verb!eps! set to values close to zero:

\begin{Verbatim}[fontsize=\small]
double sum = 0.0;
for(int i = 0; i < num_elems; i++)
    sum += x[i] * in_range(ref[i] - ref_value, -eps, eps);
return sum;
\end{Verbatim}

For brevity, we assume that the attribute value allows us to uniquely select the desired neighbor.
If the uniqueness is not guaranteed by the model, multiple reference attributes can be supplied.
For instance, in the traffic model presented in Section~\ref{sec:model_implementation}, the vehicle ahead is selected based jointly on its position and lane, which due to the model behavior is sufficient to guarantee uniqueness.

An important downside of this interaction partner selection method is its overhead, as each agent traverses all other agents.
In Sections~\ref{sec:model_implementation} and~\ref{sec:experiments}, we present opportunities for performance improvements and evaluate the overhead of different model variants.

\subsection{Time-Dependent Behavior}
\begin{figure}[b]
\vspace{-0.45cm}
\centering
\includegraphics[width=0.49\textwidth]{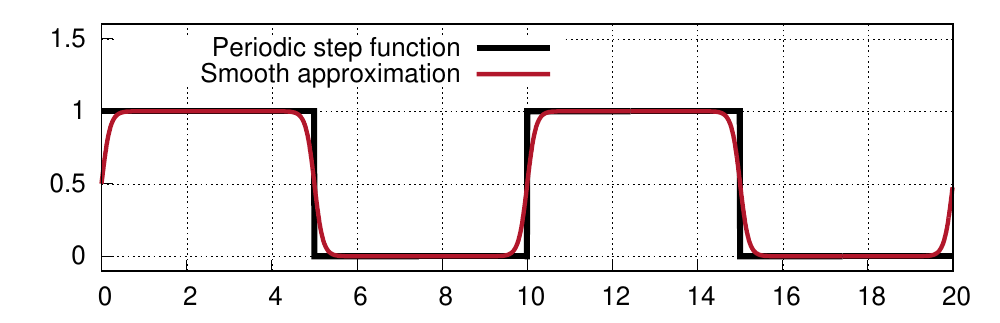}
\vspace{-0.5cm}
\caption{Smooth approximation of a periodic step function.}
\label{fig:periodic_step_function}
\end{figure}

Agents may dynamically alter their behavior throughout the simulation.
If the relation between time and behavior is not input-dependent, it can be viewed as a static part of the model logic and does not require further consideration with respect to differentiability.
An example are actions triggered every $n$ time steps.

Input-dependent periodic behavior can be expressed using the smooth periodic step function $\text{l}_k(\sin(\frac{{\pi}t}{p}))$ shown in Figure~\ref{fig:periodic_step_function}, where $t$ is the current time and $p$ is the period.
The action itself is then triggered through \emph{smooth branching}, relying on the smoothed periodic step function to mask the action if it is currently inactive.

Actions that should only occur once can be scheduled using \emph{smooth timers}: a timer variable $v_t$ is initialized to the desired delta in time and decremented at each time step, where $\text{l}_k(-v_t, 0)$ serves as a condition for \emph{smooth branching}.
When the timer is not needed, $v_t$ can be set to a positive value larger than the simulation end time.

\subsection{Stochasticity}
For the purpose of differentiation, pseudo-random numbers (PRNs) drawn based on a given seed throughout a simulation run can be regarded as constants, even if the numbers are drawn at runtime.
To estimate the gradient of a stochastic model at a given parameter combination, the gradients obtained from runs with different seeds can be aggregated.
While simple averaging can yield biased gradient estimates, recent results by Eckman and Henderson indicate that such estimates can still steer the optimization near local optima~\cite{eckman2020biased}.

Importantly, even if PRNs are treated as constants, the effects of subsequent operations on PRNs are still captured by the gradient.
An example is inverse transform sampling, which transforms uniformly distributed PRNs to a target distribution.
In the epidemics model of Section~\ref{sec:model_implementation}, the rate of the target exponential distribution is affected by the simulation input.
In that situation, the sensitivity of the exponential variate to changes in the input is captured by the gradient.
A more sophisticated treatment of stochasticity, e.g., by operating directly on distributions~\cite{chaudhuri2010smooth}, is left for future work.

\section{Model Implementation}
\label{sec:model_implementation}

We prepared four models for automatic differentiation, three of which represent vehicle traffic controlled by traffic lights.
The first variant allows gradient information to propagate through all model elements, but is limited in its scalability.
By restricting differentiability to aspects relevant to our use case of simulation-based optimization, two further variants are able to scale to road networks populated by thousands of vehicles.
Finally, we consider an agent-based formulation of the classical susceptible-infected-recovered model extended by movement on a graph.

\subsection{Microscopic Traffic Simulation}
\label{sec:model_implementation:subsec:microscopic_traffic_simulation}

The traffic simulations rely on model classes encountered in common academic and commercial traffic simulators such as SUMO~\cite{behrisch2011sumo} and VISSIM~\cite{fellendorf2010microscopic}.
The agents' longitudinal movement is governed by the Intelligent Driver Model (IDM)~\cite{Treiber2000}.
IDM defines a vehicle's acceleration by the following ordinary differential equation:
$$\frac{dv}{dt} = a_0\left(1 - \left(\frac{v}{v_d}\right)^\delta - \left(\frac{s_0 + vT + {(v\Delta}v)/(2\sqrt{a_0b_0})}{{\Delta}p}\right)^2\right)$$
Here, $a_0$ is the maximum acceleration, $v$ is the current velocity, $v_d$ is the target velocity, $s_0$ is the minimum desired distance to the vehicle ahead, $b_0$ is the maximum deceleration and ${\Delta}p$ and ${\Delta}v$ are the position and velocity deltas to the leading vehicle.
The tuning parameter $\delta$ is typically set to 4~\cite{Treiber2000}.
In time-driven microscopic traffic simulations, each vehicle determines an acceleration value for the next time step from $t$ to $t + \tau$ based on the vehicle states at $t$.
From the acceleration, the new velocity and position is determined.

For lane changing, we rely on a simplified model similar to MOBIL~\cite{kesting2007general}: every $n$ time steps,  the vehicles determine the projected increase in clearance observed after a hypothetical lane change to the left or right lane, if any.
If the clearance increase is beyond a configurable threshold, an instantaneous lane change is carried out.

The traffic is controlled by traffic lights with static or dynamic timing as described below.
Overall, we obtain hybrid models composed of an originally continuous car-following behavior, which is discretized through numerical integration, and the purely discrete transitions in the lane changing behavior and traffic light control.

\subsubsection{Single Multi-Lane Road}
The purpose of this initial model variant is to explore the viability of implementing a fully differentiable model, i.e., one in which the computed gradients capture the behavior of all model elements, and to study the computed gradients when varying the input parameters.
While we hope for the model description to be instructive, we will see that for practical applications, it seems preferable to limit the incurred overhead by restricting the differentiability to selected model elements.

We first consider the car-following model IDM, which in a time-driven formulation directly relates the new acceleration of a vehicle to its leader's current state, making automatic differentiation of the acceleration update itself straightforward.
A challenge lies in determining the leader: in a typical implementation, the vehicles located on a lane are stored in a sorted fashion, rendering the leader selection trivial.
However, differentiable sorting is a non-trivial problem currently under active research~\cite{cuturi2019differentiable,grover2019stochastic}.
Thus, given an unordered list of vehicles, we determine a vehicle's leader by iterating across all vehicles and determining the vehicle with the minimum positive position delta.
Since we require both the position and velocity of the leader, we arrive at a two-step process: first, we determine the leader's position by repeatedly applying \emph{smooth minimum} as described in Section~\ref{sec:differentiable_agent-based_simulation:subsec:selection_of_interaction_partners}, masking negative position deltas using \emph{smooth threshold}.
Finally, \emph{select by attribute} determines the leader's velocity based on its position and lane.

Lane changing decisions are made periodically by iterating across all agents and determining the lane with the largest forward clearance using \emph{smooth maximum}, after which \emph{smooth threshold} is applied to determine whether the clearance justifies a lane change.

A traffic light spanning all lanes is positioned on the road, alternating between green and red phases of equal duration using \emph{periodic step function}.
The vehicles brake at red lights according to IDM given a zero-velocity leader.
This is achieved using \emph{smooth branching} on three conditions: the light is red, the light is still ahead, and the light is closer than the leading vehicle, if any.

While this fully differentiable model variant is operational, it is prone to produce artifacts.
For instance, when a vehicle advances past the end of a road $l$ meters in length, its position is reset to the beginning of the road using the \emph{smooth threshold} building block.
If the vehicle position happens to be exactly $l$ meters, the argument to the logistic function is 0, yielding a new vehicle position of $\frac{l}{2}$m/s instead of the desired $0$m/s.
The probability for such artifacts to occur can be controlled by adjusting the slope of the logistic function (cf.~Section~\ref{sec:differentiable_agent-based_simulation}) at the cost of an increase of the magnitudes of the derivatives around 0, and a decrease everywhere else.

Given these considerations, the model variants described below follow a more pragmatic approach that restricts the differentiability to those model aspects for which we expect gradient information to directly benefit our use case of simulation-based optimization.


\subsubsection{Grid Network with Static Signal Timings}

\begin{figure}[b]
\centering
\includegraphics[width=0.40\textwidth]{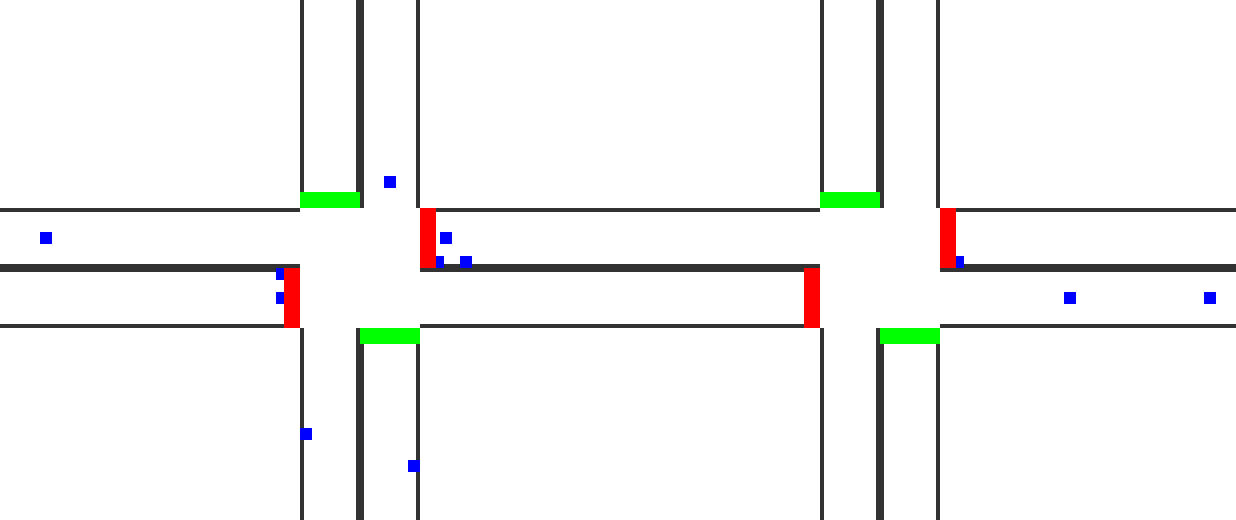}
\caption{A section of the grid scenario: the vehicles traverse a grid of light-controlled intersections. At the network boundaries, vehicle positions wrap around.}
\label{fig:grid_network}
\end{figure}

In this model variant, the vehicles traverse a grid-shaped network of multi-lane roads (cf.~Figure~\ref{fig:grid_network}) connected at the grid boundaries to form a torus.
At each intersection, traffic lights are placed at the incoming roads.
The timings of the light phases at the intersections are given as offsets in logical time, which form the simulation input.
When encountering an intersection, the vehicles turn left or right or advance to the road ahead based on configurable probabilities.
While the vehicles' behavior with regard to their acceleration, lane changing, and the traffic lights is identical to the previous model variant, the implementation follows a different approach: our ultimate objective is to maximize the overall vehicle progress by adjusting the traffic light timings.
Hence, the key model elements for which we aim to extract gradient information are the light control and the longitudinal movement of the vehicles.
The specific differences to the previous model pertain to determining a vehicle's leader, the lane changing, and the advancement to adjacent roads.
These aspects follow their natural implementations based on storing and updating the vehicle states in per-lane arrays sorted by position and are hence not captured in the computed gradients.
As an example, suppose a slight change in the simulation input would cause a new lane change, which in turn would affect the simulation output.
As the model does not offer differentiability across lane changes, the gradient would not reflect this possibility.

Aside from the performance benefits of this approach (cf.~Section~\ref{sec:experiments}), the more natural implementation suggests that integrating automatic differentiation capabilities in an existing traffic simulator could be possible without excessive development efforts.

In Section~\ref{sec:experiments:subsec:grid_network_with_static_signal_timings}, this model variant serves as an example for simulation\hyp{}based optimization.
One input parameter per intersection represents the light phase offset and will be adjusted to maximize the vehicles' progress.
To limit the input dimension, existing work considers only small numbers of intersections (e.g.,~\cite{dabiri2016arterial}) or reduces the model detail from an individual-based view as used in our work to the mesoscopic or macroscopic level (e.g.,~\cite{zhang2013robust}).

\subsubsection{Grid Network with Neural Network-Controlled Signals}
\begin{figure*}[t]
\centering
\includegraphics[width=0.95\textwidth]{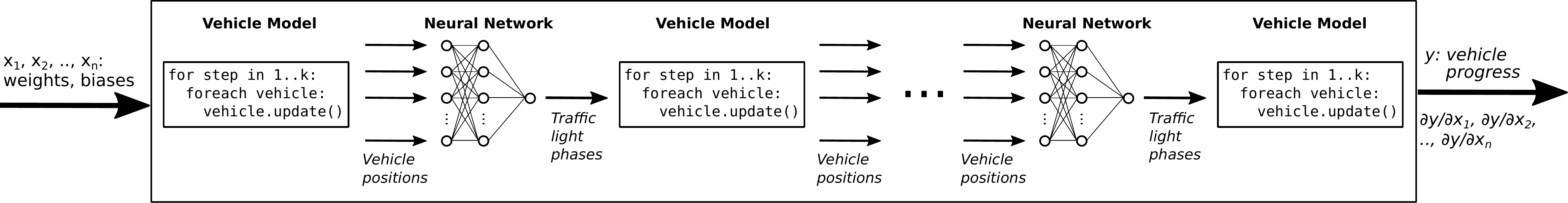}
\caption{Neural network-controlled traffic lights embedded in a traffic simulation. As the neural network is part of the differentiable model logic, training can rely on the partial derivatives w.r.t.~the network coefficients returned by the simulation.}
\label{fig:sim_nn}
\vspace{-0.25cm}
\end{figure*}

In this model variant, we substitute the traffic light control based on time offsets with a dynamic control using a neural network.
The setup is illustrated in Figure~\ref{fig:sim_nn}: the neural network is invoked periodically as part of the model logic.
At each decision point, the current positions of all vehicles in the simulation are provided as input to the neural network, its output being the new traffic light phases (red or green) for the horizontal roads at each intersection, the vertical roads being assigned the opposite phase.
Since the neural network is implemented as part of the model logic, the gradients extracted through automatic differentiation directly reflect the sensitivity of the vehicles' movement to the neural network's coefficients, enabling gradient-based training in order to optimize the traffic flow.
This is in contrast to reinforcement learning approaches, which typically operate under the assumption that a system model is not available and that the effect of the trained entity's actions must be explored by observing the resulting states and ``rewards'' throughout repeated simulation runs~\cite{kaelbling1996reinforcement}.
In our case, each simulation run returns not only an overall reward in the form of the vehicles' progress, but also derivatives that guide the optimization towards a locally optimal traffic light response.
A small number of existing works rely on automatic differentiation to train neural networks in the context of purely continuous control applications~\cite{baydin2017automatic}.
We are not aware of any work that relied on automatic differentiation to directly train neural networks embedded in the model logic of agent-based simulations or traffic simulations.

The neural network follows a fully connected feed-forward architecture: there are 5 input neurons for each lane in the road network, the input being the sorted positions of the 5 vehicles closest to the intersection.
There is a single hidden layer comprised of $h$ neurons.
The output layer is comprised of one neuron per intersection, which yield the new traffic light states.
Given $i$ intersections, 4 incoming roads per intersection, and 3 lanes per road, the architecture results in $(60i + 1)h + (h + 1)i$ coefficients to be adjusted.
All neurons rely on the hyperbolic tangent function for activation.
The traffic light states returned by the neural network are floating point numbers translated to green or red phases using \emph{smooth threshold}, a positive value representing a green light at a horizontal road.

\subsection{Epidemics Model on a Graph}
\label{sec:model_implementation:subsec:epidemics_model_on_a_graph}
The final model follows Macal's agent-based formulation~\cite{macal2010agent} of the well-known Susceptible-Infected-Recovered model~\cite{kermack1927contribution}, which imitates the spread of an infection.
We extend this model by random movement on a social graph.
The model serves to illustrate the viability of automatic differentiation given purely discrete agent states and under strong dependence on stochastic model elements.

An arbitrary number of agents is situated on each node of a static graph.
Initially, each agent is either in the ``susceptible'' or ``infected'' state, the probability being a model input.
At each time step, each agent $a$ acts as follows: if $a$ is susceptible, each infected agent $a' \ne a$ at the current location infects $a$ with a per-location probability given as an input.
If $a$ is newly infected, the delay to the transition to the ``recovered'' state is drawn from an exponential distribution, the rate being another input.
Finally, the agent moves to a neighboring graph node chosen uniformly at random.

For a given seed value, the agents change their locations according to predetermined trajectories.
As the sequence of visited locations is thus fixed, the overhead of differentiable neighbor search can be avoided by gathering each agent's neighbors from an array updated each time step.
The key remaining model aspects are constructed from differentiable building blocks: infections are handled by supplying uniformly distributed random variates and the infection probabilities as input to \emph{smooth branching}.
Recovery is an instance of the \emph{smooth timer} building block.
Using as input a uniform random variate and the simulation input specifying the recovery rate, we determine the concrete delay until the agent recovers using inverse transform sampling.
As the gradient information propagates through the uniform-to-exponential transformation, the computed gradients capture the sensitivity of the simulation output to the configured recovery rate.
At each time step, the \emph{smooth branching} building block is applied to carry out the transition to the ``recovered'' state once the recovery delay has expired.

\section{Experiments}
\label{sec:experiments}

Our experiments are intended to answer the research question:\linebreak
\emph{``Can gradient-based simulation-based optimization using the presented differentiable models outperform gradient-free methods?''}

To achieve a benefit during optimization, the fidelity of the differentiable model must be sufficient so that the quality of identified solutions carries over to the non-differentiable reference model.
Further, the execution time overhead of the simulation must be small enough not to outweigh potential improvements in convergence speed.
Thus, in the remainder of the section, we evaluate the fidelity and overhead of the differentiable model variants.
The overall benefit of gradient-based over gradient-free optimization is evaluated in a number of simulation-based optimization experiments.

The experiments were conducted on two machines each equipped with two 16-core Intel Xeon CPU E5-2683v4 and 256GiB of RAM, running CentOS Linux 7.9.2009.
The automatic differentiation relied on Adept 1.1~\cite{hogan2014fast}.
For optimization, we used ensmallen 2.15.1~\cite{bhardwaj2018ensmallen}.

\subsection{Single Multi-Lane Road}
\label{sec:experiments:subsec:single_multi-lane_road}

We first study the deviation of the results generated by the fully differentiable model as compared to a non-differentiable reference implementation.
The road has three lanes 250m in length.
A traffic light is positioned at 100m, with an overall period of 10s, divided into green and red phases of 5s each.
The speed limit is set to 50km/h.
Lane changes may occur every 2.5s, requiring a minimum clearance increase of 10m.
The IDM parameters defining the maximum acceleration and deceleration are both set to 2m/s.
Where not otherwise noted, the same parameters are used in the microscopic traffic simulation experiments presented below.
The time step size $\tau$ is set to 0.1s.
Initially, we position the vehicle at different lanes in non-zero increments of 40m from the beginning of the road.

\begin{figure}[b]
\squeeze
\vspace{-0.25cm}
\centering
\begin{minipage}[b]{.47\textwidth}
    \includegraphics[width=\textwidth]{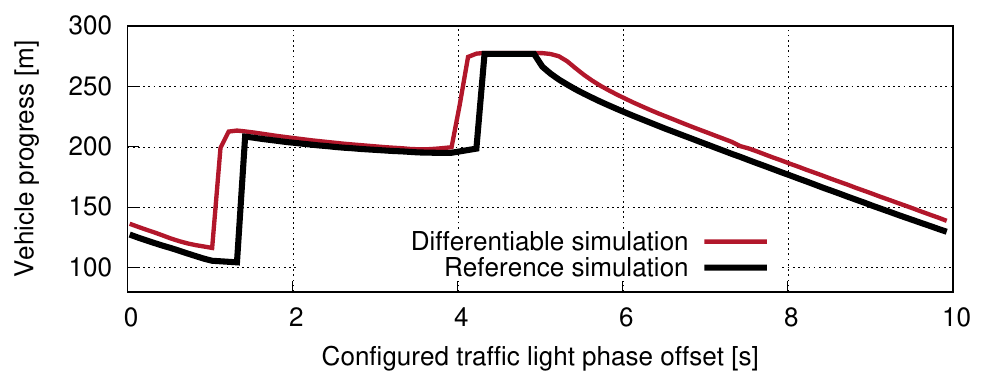}
	\subcaption{Vehicle progress.}
    \label{fig:traffic_sim_single_road_y_2_veh}
\vspace{-0.02cm}
\end{minipage}
\begin{minipage}[b]{.47\textwidth}
    \includegraphics[width=\textwidth]{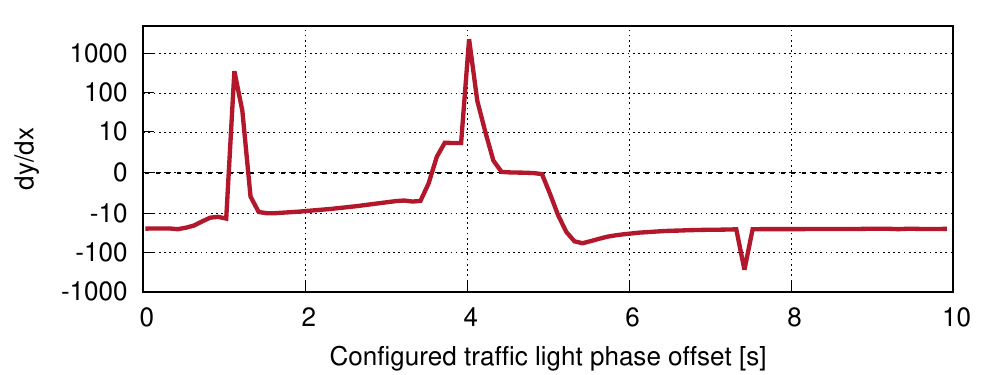}
	\subcaption{Derivative of the vehicle progress w.r.t.~the phase offset.}
    \label{fig:traffic_sim_single_road_dydx_2_veh}
\end{minipage}
\vspace{-0.22cm}
\caption{Overall vehicle progress and its derivative in the single-road scenario with two vehicles.}
\label{fig:traffic_sim_single_road_2_veh}
\end{figure}

\begin{figure}[t]
\centering
\includegraphics[width=0.49\textwidth]{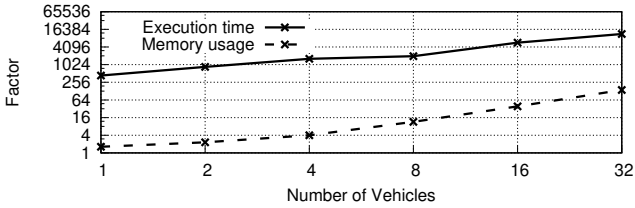}
\caption{Performance of the fully differentiable traffic simulation relative to a non-differentiable reference simulation. The overhead increases strongly with the vehicle count.}
\squeeze
\label{fig:traffic_sim_single_road_factors}
\end{figure}

Figure~\ref{fig:traffic_sim_single_road_y_2_veh} compares the vehicle progress in meters throughout a simulation involving two vehicles spanning 10s of logical time between the differentiable simulation and the non\hyp{}differentiable reference.
The x axis shows the simulation input, which is the time at which the traffic light first changes to red.
We show results with the slope parameter of the logistic function set to 32.

The sharp increases in the vehicle progress around 1s and 4s reflect situations where a vehicle passes the light just before it changes, instead of braking sharply.
The curve for the differentiable simulation slightly deviates from the reference, the reason being the smoothing of the traffic light control, which delays the braking and acceleration when the light changes.

Figure~\ref{fig:traffic_sim_single_road_dydx_2_veh} shows the derivative of the simulation output w.r.t.~the input parameter determined by automatic differentiation.
The derivative expresses the sensitivity of the vehicle progress to changes in the traffic light offset.
We can see that overall, the derivative follows the slope of the simulation output curve, sharp increases in the simulation output being reflected by spikes in the derivative.
However, an additional negative spike at an offset of around 7.5s illustrates that the derivative only represents the slope in a given point of the simulation input and is thus sensitive to implementation artifacts.
Artifacts of this type occur when a real value that represents a boolean or integer deviates too far from its reference value.
In Figure~\ref{fig:traffic_sim_single_road_y_2_veh}, the resulting miniscule ``dent'' in the simulation response, which is not present in the reference simulation, translates to a large derivative.
This effect becomes more pronounced when increasing the slope of the logistic function.

We measured the performance of the differentiable simulation compared to the basic reference implementation.
Figure~\ref{fig:traffic_sim_single_road_factors} shows the relative wall-clock time per run and the relative memory usage.
The main overhead is induced during the simulation itself, during which the operations are recorded in preparation for the subsequent differentiation step.
The contribution of the differentiation at 32 vehicles was about 170s of a total of 789s, or about 22\%.

Given the issue of model artifacts and the enormous overhead of this fully differentiable model, the models evaluated in the following restrict the differentiable aspects to the traffic light control and the vehicles' forward movement.
In Section~\ref{sec:limitations_and_research_directions}, we discuss further options to reduce the overhead while maintaining the differentiability of the main model components.

\subsection{Grid Network with Static Signal Timings}
\label{sec:experiments:subsec:grid_network_with_static_signal_timings}
We now evaluate the traffic model where we restricted the differentiable aspects to the traffic light control and the vehicles' longitudinal movement.
The grid network used in the experiments is comprised of three-lane roads 100m in length, with a speed limit of 35km/h.
At an intersection, a vehicle advances to the road on the left, right, or straight ahead with probabilities 0.05, 0.05, and 0.9.
The slope parameter of the logistic function was set to 32.

\begin{figure}[t]
\centering
\includegraphics[width=0.49\textwidth]{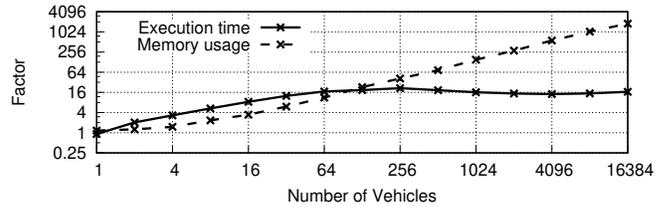}
\caption{Performance of the traffic simulation on a grid network relative to a non-differentiable reference implementation. The overhead is substantially smaller compared to the fully differentiable model (cf.~Fig.~\ref{fig:traffic_sim_grid_factors}).}
\label{fig:traffic_sim_grid_factors}
\end{figure}

Figure~\ref{fig:traffic_sim_grid_factors} shows the performance comparison to the non\hyp{}differentiable reference simulation on a grid comprised of 50x50 intersections spanning 180s of logical time.
The traffic lights alternate between red and green phases of 10s each.
While there is still a substantial overhead in memory usage and execution time, the overhead is significantly lower than with the fully differentiable model.
In particular, the execution time factor is now only weakly affected by the number of vehicles.
In contrast to the execution time, the relative memory consumption still increases with the number of vehicles as the results of more and more intermediate computational operations have to be stored to permit the subsequent gradient computation.
Still, with an absolute execution time of 2.7s and a memory usage of 1.6GiB with 1\,024 vehicles, and 47.2s using 23.2GiB with 16\,384 vehicles, we consider this model to be sufficiently scalable to cover scenarios of practical relevance.

We now turn to the comparison of the gradient-based and gradient-free simulation-based optimization.
Our goal is to maximize the vehicle progress by adjusting the traffic light timings.
The following gradient-free optimization methods are employed: Differential Evolution (DE)~\cite{storn1995simple}, Conventional Neural Evolution (CNE)~\cite{montana1989training}, and Simulated Annealing (SA)~\cite{kirkpatrick1983optimization}
To include a representative of methods based on gradient estimations, we also show results using Stochastic Perturbation Stochastic Approximation (SPSA)~\cite{spall1992multivariate}.
The optimization using differentiable simulation relies on the following gradient-based methods: Stochastic Gradient Descent (SGD), Adaptive Moment Estimation (Adam)~\cite{kingma2014adam}, and Nadam~\cite{dozat2016incorporating}.

The optimizers are configured using varying numbers of parameters.
To limit the number of optimization runs for evaluation, we adjust only two parameters and use the default values configured in the ensmallen library for the remaining parameters: 1.~The step size (or its equivalent) is set to the values $\{10^{-3}, 10^{-2}, .., 10^{0}\}$, and 2.~for DE and CNE, which combine the current best solutions only after completing a so-called generation of runs, we reduce the generation size from its default of 500 to 50.
Each optimization process starts from the same randomly initialized parameter combination.
For each optimizer, we show the improvement over the initial parameter combination for the step size that achieved the maximum value within the time budget of 72 hours.

\begin{figure}[t]
\centering
\begin{minipage}[b]{.47\textwidth}
    \includegraphics[width=\textwidth]{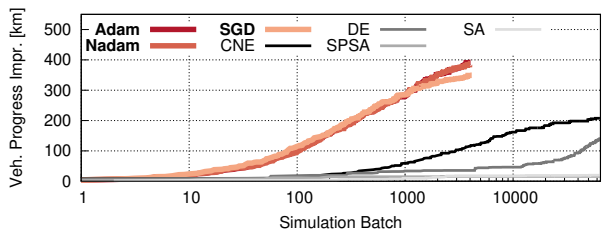}
    \subcaption{Best observed solution across batches of 10 runs each.}
    \label{fig:opt_traffic_sim_grid_batches_50x50_20_10}
\vspace{0.4cm}
\end{minipage}
\begin{minipage}[b]{.47\textwidth}
    \includegraphics[width=\textwidth]{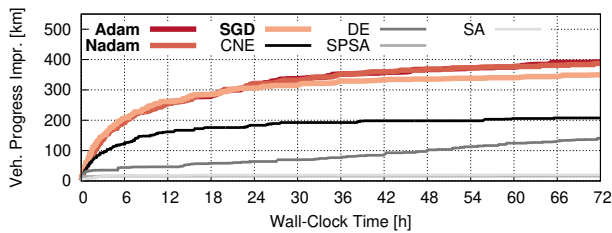}
    \subcaption{Best observed solution across wall-clock time.}
    \label{fig:opt_traffic_sim_grid_time_50x50_20_10}
\end{minipage}
	\caption{Optimization progress for the 50x50 grid scenario using static traffic light control with a period of 20s.}
\end{figure}

Figure~\ref{fig:opt_traffic_sim_grid_batches_50x50_20_10} shows the optimization progress as the improvement over the initial random parametrization for a 50x50 grid populated with 2\,500 vehicles, with an overall traffic light control period of 20s.
The total number of parameters to be adjusted is 2\,500.
Each point in the parameter space is evaluated by executing a batch of 10 simulation runs, averaging the returned output values, and, for the differentiable simulation, the gradients.
To avoid an excessive impact of large individual derivatives, we employ gradient clipping~\cite{pascanu2013difficulty}, restricting the derivatives to the arbitrarily chosen interval $[-10, 10]$.
The same initial vehicle positions are used in all runs, randomizing only the turns at intersections, to introduce sufficient regularity in the traffic patterns to permit an optimization of the traffic light timings.
Due to the overhead of the differentiable simulation, the gradient-based methods executed only about 4\,000 simulation batches within the time budget, compared to about 64\,000 with the gradient-free methods.
However, the optimization progress per batch is immensely faster than with the gradient-free methods.
For instance, after 100 batches, all gradient-based methods achieve an improvement of about 100km, whereas the improvement achieved by any of the gradient-free methods was still below 20km.

Since practical applications are concerned with the solution quality achieved within a given time or compute budget, Figure~\ref{fig:opt_traffic_sim_grid_time_50x50_20_10} shows the optimization progress across wall-clock time.
Due to the faster execution of the non-differentiable simulation, the initial benefit of the gradient-based optimization is somewhat reduced.
Still, the best solution quality up to any given point in time is still vastly superior with the gradient-based methods.

A gradient-based optimization could also be carried out using finite differences based on the non-differentiable simulation, which is faster than its differentiable counterpart by a factor of about 16.
However, given the 2\,500 simulation inputs, 2\,501 batches would need to be executed in each point to obtain a gradient estimate, requiring roughly 2.7h of wall-clock time per point.
Since this would allow for an exploration of only about 25 parameter combinations within the time budget of 72h, we abstained from implementing the gradient-based optimization using finite differences.

To show the validity of the comparison between the gradient-based and gradient-free optimization results, we used the highest-quality solution returned by Adam as input to the non-differentiable reference simulation, executing 100 runs.
The mean overall vehicle progress including the progress in the initial solution, with 95\% confidence intervals was 3\,467.2 $\pm$ 0.8km in the differentiable simulation and 3\,375.1 $\pm$ 0.9km in the non-differentiable simulation.
For comparison, the best solution found using CNE translated to only 3\,175.2 $\pm$ 0.9km of vehicle progress.

\begin{figure}[t]
    \centering
    \includegraphics[width=.47\textwidth]{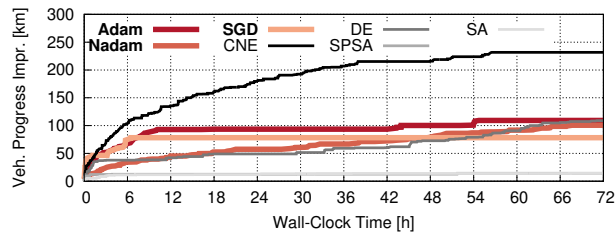}
	\caption{Optimization progress with 50x50 intersections and a traffic light control period of 40s.}
    \label{fig:opt_traffic_sim_grid_time_50x50_40_20}
    \squeeze
\end{figure}

We repeated the experiment after increasing the traffic light period from 20s to 40s.
In Figure~\ref{fig:opt_traffic_sim_grid_time_50x50_40_20}, we see that in this configuration, the gradient-based methods are outperformed by CNE, achieving a similar solution quality as DE.
A likely reason is given by the \emph{periodic step function}: with a longer light period, there is a larger probability of generating very small gradients, which pose a challenge to the gradient-based methods~\cite{hochreiter1998vanishing} (cf.~Section~\ref{sec:limitations_and_research_directions}).

\begin{figure}[b]
    \centering
\centering
\begin{minipage}[b]{.47\textwidth}
    \includegraphics[width=\textwidth]{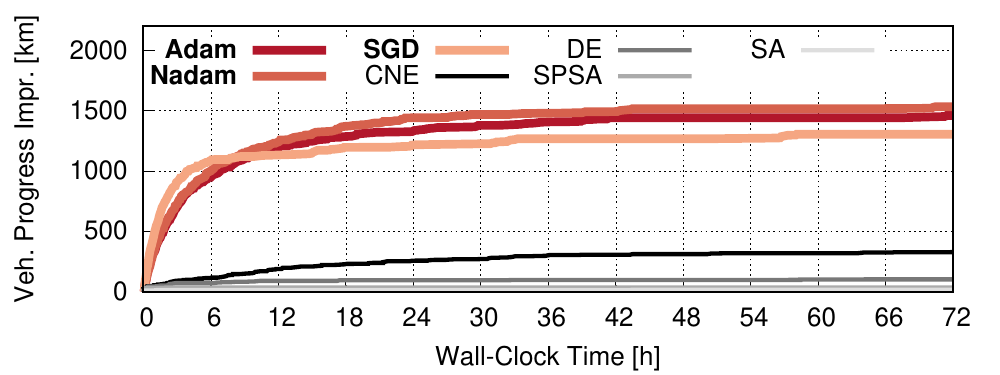}
    \squeeze
	\subcaption{Traffic light control period of 20s.}
    \label{fig:opt_traffic_sim_grid_time_100x100_20_10}
\end{minipage}
\begin{minipage}[b]{.47\textwidth}
    \includegraphics[width=\textwidth]{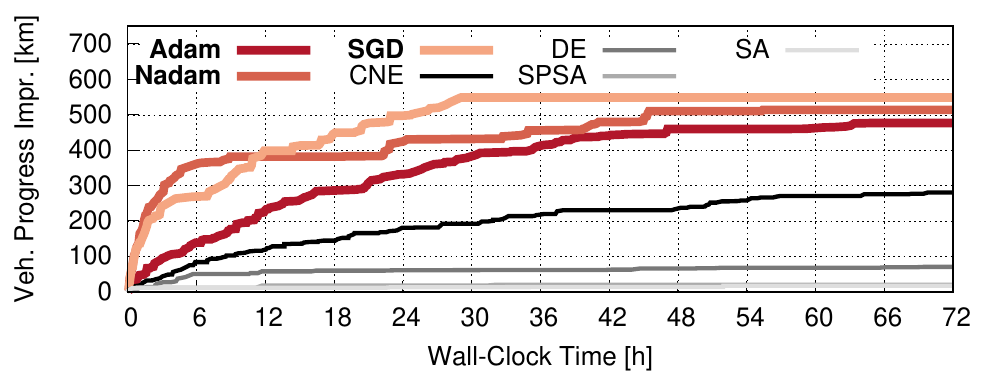}
	\subcaption{Traffic light control period of 40s.}
\end{minipage}
    \squeeze
    \caption{Optimization progress, 100x100 intersections.}
    \label{fig:opt_traffic_sim_grid_time_100x100_40_20}
\end{figure}

The experiment was repeated with a network of 100x100 intersections and 10\,000 vehicles, which increases the number of parameters to 10\,000 while maintaining the same vehicle density.
Figures~\ref{fig:opt_traffic_sim_grid_time_100x100_20_10} and~\ref{fig:opt_traffic_sim_grid_time_100x100_40_20} show that as a result, the advantage of the gradient-methods becomes more pronounced, with consistently and vastly better solution quality compared to the gradient-free methods.

\subsection{Grid Network with Neural Network-Controlled Signals}
\label{sec:experiments:subsec:grid_network_with_nn-controlled_signals}

A similar optimization experiment as above was carried out on a 5x5 grid populated by 200 vehicles, introducing dynamic traffic light control by a neural network.
Given that each decision of the traffic light control remains in effect for 20s, an optimal policy would consider not only the vehicle positions at each intersection, but would also include the positions and traffic light phase at the neighboring intersections.
The number of neurons in the hidden layer was set to 60, resulting in 91\,585 neural network coefficients forming the simulation input.
All optimization methods started from the same initial parameter combination drawn from a standard normal distribution.
After preliminary experiments, we accelerated the optimization progress by randomizing the initial vehicle positions for each run and by modifying the simulation output to be the minimum progress among the vehicles instead of the sum progress.

\begin{figure}[t]
    \centering
    \includegraphics[width=0.49\textwidth]{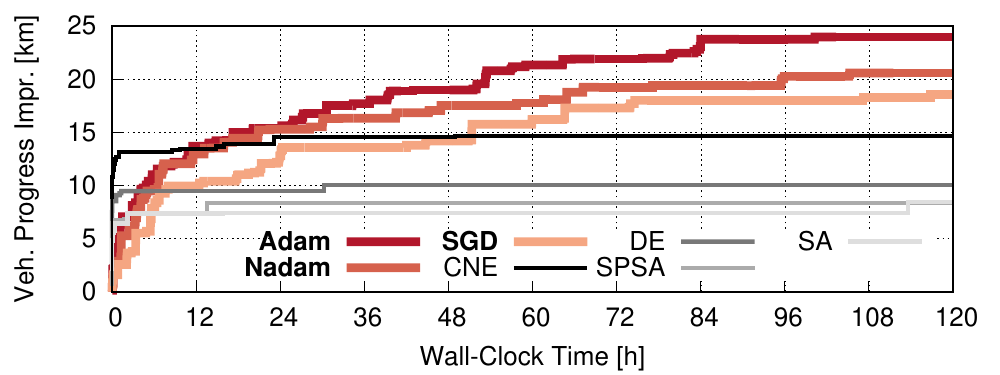}
    \vspace{-0.2cm}
	\caption{Optimization progress for the grid traffic scenario with neural network-controlled traffic lights.}
    \label{fig:opt_traffic_sim_grid_nn_time}
    \vspace{-0.3cm}
\end{figure}

Figure~\ref{fig:opt_traffic_sim_grid_nn_time} shows that for this problem, the gradient-free methods achieve somewhat faster initial progress.
However, beyond about 12 hours, Adam and Nadam outperform all gradient-free methods.
At the end of the time budget of 120h, the highest overall vehicle progress achieved by Adam, as measured in the non-differentiable simulation, was 47.8 $\pm$ 0.7km.
The best result among the gradient-free methods achieved by CNE was 31.3 $\pm$ 0.9km.
Given 91\,585 inputs, the cost of the finite differences method is again prohibitive.

\subsection{Epidemics Model on a Graph}
\label{sec:experiments:subsec:epidemics_model_on_a_graph}

\begin{figure}[t]
\centering
\vspace{0.2cm}
\includegraphics[width=0.44\textwidth]{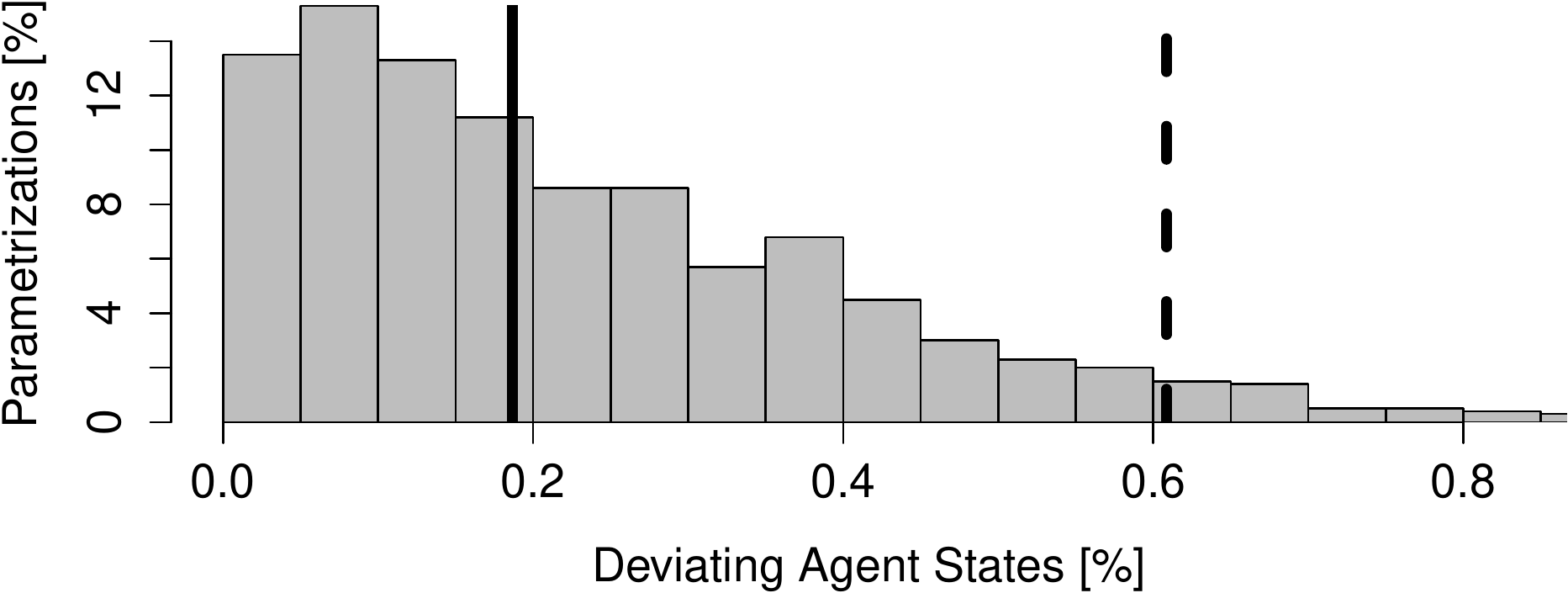}
\vspace{-0.2cm}
\caption{Comparison of results from the differentiable and non-differentiable epidemics models. The vertical and dashed lines indicate the median and 95\%-quantile.}
\label{fig:epidemics_verification}
\vspace{-.2cm}
\end{figure}

For the Susceptible-Infected-Recovered model, we are particularly interested in the fidelity since the differentiable variant must represent the originally purely discrete agent states as real numbers.
We executed 100 simulations each for 1\,000 parametrizations of a scenario populated by 1\,000 agents moving across a random geometric graph with 500 nodes and an average degree of 5, each run spanning 10 time steps.
The per-location infection rate coefficients and the initial infection probability were drawn from $U(0, 0.1)$.
The recovery rate was drawn from $U(0, 0.01)$.
The slope of the logistic function was set to 32.
Figure~\ref{fig:epidemics_verification} shows a histogram of the percentage of agents attributed to a different state than in the non-differentiable reference runs.
The median deviation amounts to 0.19\% of the agents, and the 95\% and 99\% quantiles are 0.61\% and 0.83\%, which indicates that the differentiable model closely represents the reference model.

We also conducted an optimization experiment in which we calibrated a simulation of 10\,000 agents on a graph of 5\,000 nodes to a state trajectory across 10 steps of a randomized reference run.
Similar recent work operates on a surrogate for an original agent-based model~\cite{anirudh2020accurate}.
We summarize our results briefly: CNE, DE, and the gradient-based methods all achieved a good fit to the reference trajectory within the time budget of 12h.
The results using CNE, DE were somewhat superior, with about 0.7\% and 1.1\% misattributed agent states, compared to between 1.9\% and 2.2\% using the gradient-based methods.
For comparison, the solutions identified by SPSA and SA misattributed 23.0\% and 25.3\% of the states.

\section{Limitations and Research Directions}
\label{sec:limitations_and_research_directions}
Our experiments show the viability of differentiable agent-based simulation and its benefit in several simulation-based optimization problems.
Still, a number of avenues remains to be explored, the main focal points being the fidelity and performance of the differentiable models and the applicability of the approach by model domain experts and in the context of machine learning.

\textbf{Fidelity:} Most of the presented building blocks rely on approximations using the logistic function, the error being adjusted by a slope parameter.
The configuration of the slope involves a tradeoff: with steep slopes, the logistic function closely approximates a step function.
However, negative or positive arguments with sufficiently large magnitude quickly approach 0 or 1, respectively.
Similar to the vanishing gradient problem in machine learning~\cite{hochreiter1998vanishing}, the resulting small gradients may lead to a sluggish optimization.
On the other hand, with shallow slopes, arguments close to zero yield large deviations from the original step function.
An important direction for future work lies in determining model-specific error bounds based on known bounds for the building blocks (e.g.,~\cite{kyurkchiev2015sigmoid}), and on the detection and potential correction of artifacts.

\textbf{Performance:} We have seen that the overhead of the differentiable model variants is substantial, limiting their applicability for large scenarios.
One of the causes is the implementation of branching: in effect, the operations of all possible branches are executed.
While we showed that the combination of non-differentiable and differentiable model elements can limit the overhead, a key challenge lies in identifying for which model aspects gradient information is required.
For instance, in the scalable traffic model variants, the impact of variations in the simulation input on lane changes are not captured in the computed gradients.

If an optimization targets the steady-state behavior of a model, some overhead could be avoided by first executing a fast non-differentiable implementation.
Once a steady state has been reached, the simulation state is used to initialize a differentiable implementation using which the output and gradient are computed.

The memory consumption may potentially be reduced by forming so-called super nodes~\cite{margossian2019review}: first, groups of operations are identified that are repeatedly executed.
Then, by manually defining an operation that represents the contribution of an entire group to the partial derivatives, the gradient computation is simplified.
In agent-based simulations, sequences of operations executed for every agent and at every time step may be candidates for super-nodes.

Finally, the convergence speed and solution quality of simulation-based optimizations could be improved by combining gradient-free and gradient-based optimization methods.
For instance, a gradient-free approach such as a genetic algorithm could identify promising areas in the parameter space, within which a local optimum is then identified using a gradient-based method~\cite{hu2003hybridization,harada2006hybridization}.

\textbf{Applicability:} 
The models presented in Section~\ref{sec:model_implementation} were implemented manually, which, despite our relatively simple models, proved to be somewhat cumbersome and error-prone.
Automatic translations could support more convenient development processes.
Recent efforts aim to define differentiable general-purpose programming languages (e.g.~\cite{shaikhha2018efficient}).
Domain-specific languages defined in a similar vein could cater to agent-based modeling experts.

Some recent research aims at integrating differentiable programming facilities into machine learning frameworks such as PyTorch~\cite{ketkar2017introduction}.
Implementing differentiable agent-based models within such frameworks would enable a natural and efficient unification of simulation-based optimization and neural network training, making use of the frameworks' optimized GPU-based implementations of neural networks and automatic differentiation while accelerating the model execution through fine-grained many-core parallelism. 

\section{Related Work}
\label{sec:related_work}
Approximate computing techniques carry out computations at reduced fidelity, e.g., by scaling the numerical precision or by relying on neural network-based function approximation~\cite{mittal2016survey}.
Often, the intention is to reduce the computational cost, to increase resilience to errors, or to solve problems for which an exact solution is not known.
In contrast to these aims, the goal of our approximations is to allow automatic differentiation to extract gradient information.
In the context of machine learning, there is currently intensive research towards approximate differentiable algorithms for problems such as sorting~\cite{grover2019stochastic,cuturi2019differentiable,blondel2020fast} and path finding~\cite{tamar2016value}.
In future work, we plan to build on such approximate algorithms to express increasingly complex agent behavior in a differentiable manner.

Some existing works propose methods to enable the gradient-based optimization of simulations.
In Infinitesimal Perturbation Analysis (IPA)~\cite{ho1983new} and Smoothed IPA~\cite{gong1987smoothed}, gradient expressions are derived through an analysis of the given model.
While IPA can yield computations similar to those carried out by automatic differentiation, the IPA literature derives model-specific gradient estimators manually (e.g.,~\cite{howell2006simulation,chen2010perturbation,geng2012multi}), which is limited to relatively simple models.
In contrast, automatic differentiation allows gradients to be computed directly from model implementations in general-purpose programming languages.

Smooth interpretation~\cite{chaudhuri2010smooth} is a method that aims to achieve differentiability for general programs.
The program input is supplied in the form of Gaussian random variables and propagated through a symbolic execution of the program, approximating the resulting complex distributions by combinations of Gaussian distributions based on rules defined in a smoothed semantics.
The approach constitutes a potential alternative to our construction of differentiable model implementations based on a set of smooth building blocks.
An exploration of the overhead of smooth interpretation and its ability to accurately capture the logic of agent-based models is a potential avenue for future work.

Considering existing research adjacent to agent-based simulation, a recent work proposes a continuous approximation of cellular automata (CAs) to enable gradient-based search for CAs with desired properties~\cite{martin2017differentiable}.
As in IPA, expressions for the partial derivatives are determined manually.
Finally, Kreiss et al.~outlined preliminary work towards the use of automatic differentiation to calibrate Helbing's Social Force model for pedestrian dynamics~\cite{helbing1995social} against real-world data~\cite{kreiss2019automatic}.
Since the Social Force model is specified with respect to continuous time and space, it is a natural candidate for automatic differentiation.
These works share our goal of enabling gradient-based optimization, but rely on specific model properties and do not propose more general building blocks to construct differentiable agent-based simulations.

\section{Conclusions}
\label{sec:conclusions}

Simulation-based optimization of agent-based models with large numbers of inputs is usually carried out either on surrogate models, which typically abandon the individual-based level of detail of an original model, or using gradient-free methods such as genetic algorithms.
To enable direct gradient-based optimization of agent-based models, we proposed the construction of differentiable implementations using smooth building blocks, enabling an automatic computation of the partial derivatives reflecting the sensitivity of the simulation output to the inputs.

Our evaluation on the example of three variants of a road traffic model and an epidemics model was driven by the question whether gradient-based optimization using differentiable models can outperform gradient-free methods.
By constructing models from combinations of differentiable and non-differentiable model elements, we achieved sufficient performance to tackle scenarios populated by thousands of agents.
Comparing the relative solution quality and convergence speed of gradient-based and gradient-free methods in simulation-based optimization experiments, we observed that the gradient-based methods in fact achieved better results in several cases.
Particularly vast margins were observed in problems with large input dimension, which indicates that the approach could extend the reach of simulation-based optimization using agent-based models to problems that could previously only be tackled via surrogate modeling at a reduced level of detail.
As an additional benefit of the approach, we demonstrated that neural network-controlled simulation entities embedded into the differentiable model logic can efficiently be trained using gradient-based methods, with substantially superior results over gradient-free methods.

Promising research directions lie in reducing the overhead of differentiable simulations, in providing language support for expressing differentiable models in a natural way, and in model implementations targeting machine learning frameworks.

\section*{Acknowledgment}
Financial support was provided by the Deutsche Forschungsgemeinschaft (DFG) research grant UH-66/15-1 (MoSiLLDe).

\bibliographystyle{ACM-Reference-Format}
\bibliography{references}
\end{document}